\documentclass[10pt,twocolumn,letterpaper]{article}

\usepackage[pagenumbers]{cvpr} 

\usepackage{svg}
\definecolor{cvprblue}{rgb}{0.21,0.49,0.74}
\usepackage[pagebackref,breaklinks,colorlinks,allcolors=cvprblue]{hyperref}
\usepackage{tcolorbox}
\tcbuselibrary{breakable}
\usepackage{xcolor}

\newcommand{\flux}{\textsc{FLUX.1}}

\def\paperID{26} 
\def\confName{CVPR}
\def\confYear{2025}

\title{Interpreting Large Text-to-Image Diffusion Models with Dictionary Learning}

\author{Stepan Shabalin\\
Georgia Institute of Technology\\
{\tt\small sshabalin3@gatech.edu}
\and
Dmitrii Kharlapenko\\
ETH Zurich
\and
Yixiong Hao\\
Georgia Institute of Technology\\
{\tt\small yhao96@gatech.edu}
\and
Ayush Panda\\
Georgia Institute of Technology
\and
Abdur Raheem Ali\\
Trajectory Labs
\and
Arthur Conmy \thanks{Senior author as part of the MATS program.}\\
}

\begin{document}

\maketitle
\begin{abstract}
Sparse autoencoders are a promising new approach for decomposing language model activations for interpretation and control. They have been applied successfully to vision transformer image encoders and to small-scale diffusion models. Inference-Time Decomposition of Activations (ITDA) is a recently proposed variant of dictionary learning that takes the dictionary to be a set of data points from the activation distribution and reconstructs them with gradient pursuit. We apply Sparse Autoencoders (SAEs) and ITDA to a large text-to-image diffusion model, Flux 1, and consider the interpretability of embeddings of both by introducing a visual automated interpretation pipeline. We find that SAEs accurately reconstruct residual stream embeddings and beat MLP neurons on interpretability. We are able to use SAE features to steer image generation through activation addition. We find that ITDA has comparable interpretability to SAEs.
\end{abstract}
    
\section{Introduction}
\label{sec:intro}

In recent years, text-to-image model capabilities have rapidly improved \cite{carlini2023extracting}. These models develop internal representations of the physical structure of the world \cite{chen2023surfacestatisticsscenerepresentations,zhan2024generalprotocolprobelarge}. However, ongoing debates persist regarding the extent to which text-to-image models memorize training data \cite{shan2024nightshade} and how precisely their outputs can be controlled. A deeper understanding of the feature compositions of these models and the ability to intervene in their generation process would enhance artistic expression and improve transparency in how these models operate. 

In this work, we explore the interpretability of hidden representations by learning sparse decompositions of model activations, evaluating the feasibility and scalability of this approach for large text-to-image diffusion models.

Our work makes the following contributions:

\begin{enumerate}
    \item We evaluate and scale sparse autoencoders (SAEs; \citet{andrewng,cunningham,bricken2023monosemanticity}) for text-to-image diffusion models, implementing several training efficiency improvements on very large models (\Cref{sec:impl}).
    \item We assess Inference Time Decomposition of Activations (ITDA; \citet{itda}), a new dictionary learning technique, and compare its interpretability to sparse autoencoders.
    \item We demonstrate general-purpose steering of image representations using sparse autoencoder latents.
\end{enumerate}

\begin{figure}
    \centering
    \begin{subfigure}[b]{0.9\columnwidth}
        \centering
        \includegraphics[trim=0cm 0cm 3cm 0cm,clip=true,width=\columnwidth]{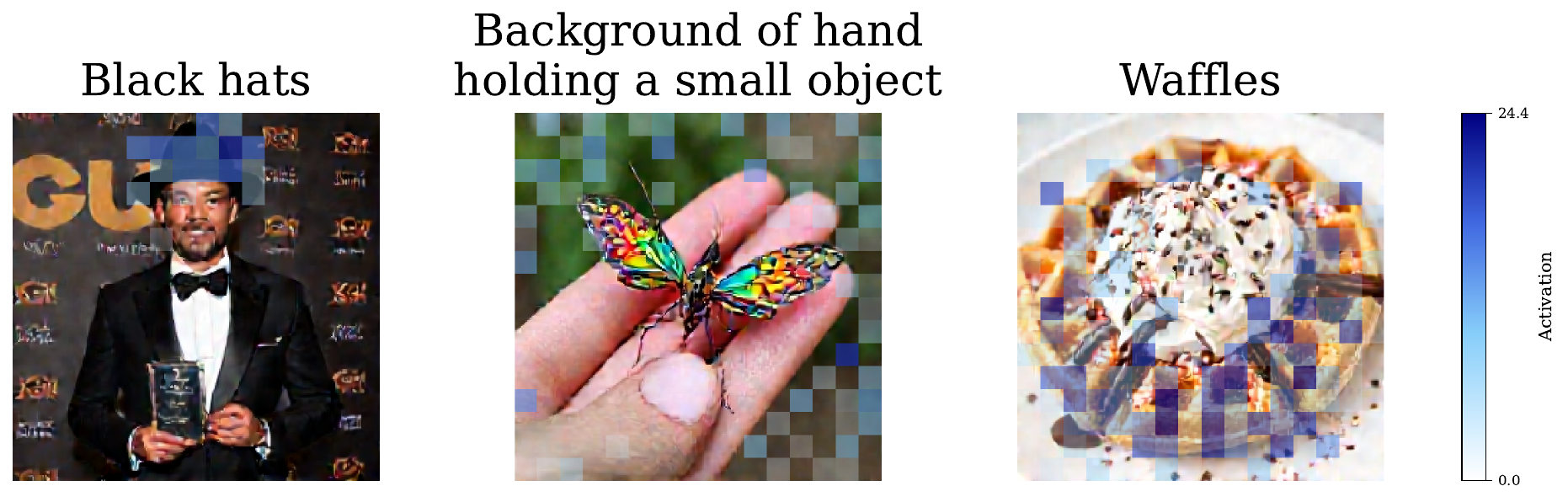}
        \caption{Maximum activating examples, LLM-generated explanations}
        \label{fig:maxacts}
    \end{subfigure}
    
    \vspace{0.5cm}
    
    \begin{subfigure}[b]{0.9\columnwidth}
        \centering
        \includegraphics[width=\columnwidth]{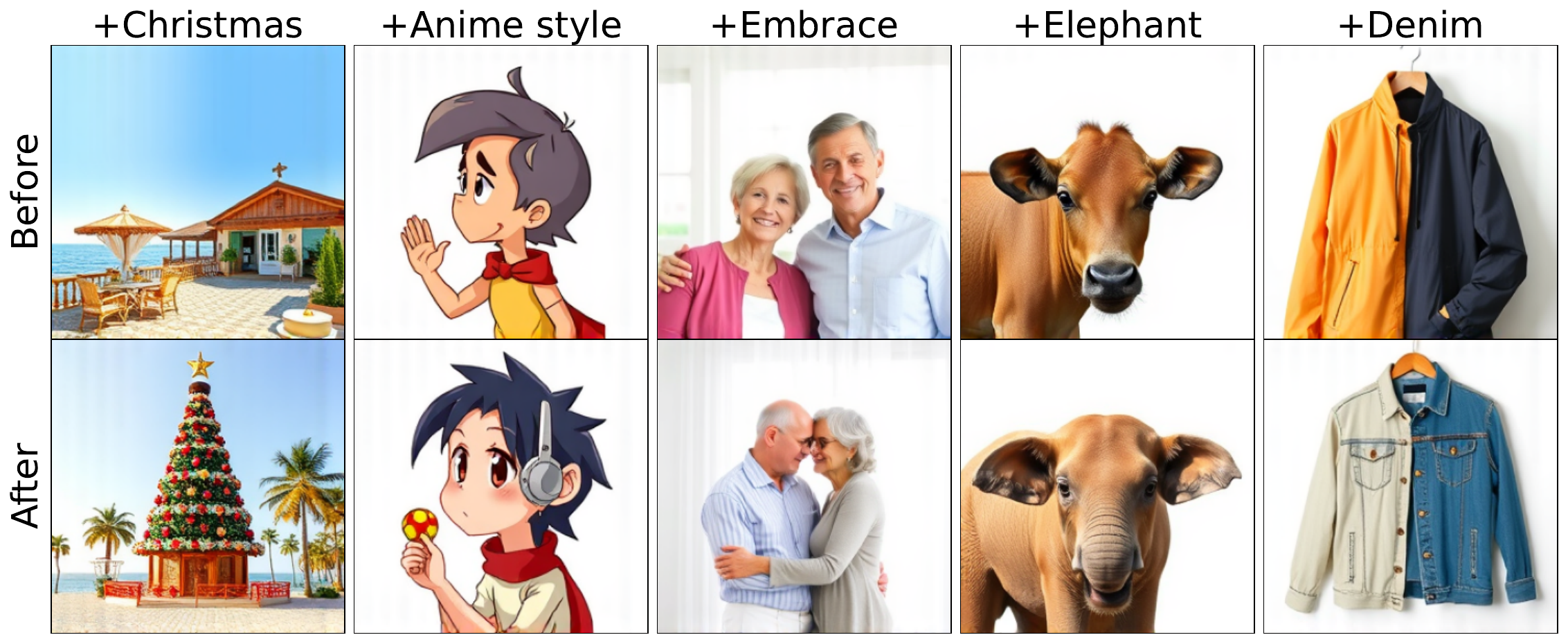}
        \caption{Steering examples}
        \label{fig:steering}
    \end{subfigure}
    
    \caption{Maximum activating examples and steering effects of some interpretable features from our \textsc{flux} SAE. See ~\Cref{more-activations} for more features.}
    \label{fig:activations}
\end{figure}

\section{Related work}
\label{sec:related}

Sparse autoencoders (SAEs, \cite{takingfeatures,bricken2023monosemanticity,huben2024sparse}) are a method for training small neural networks to resolve feature superposition \citep{elhage2022superposition} in larger networks. They work by learning a large, overcomplete decoder basis for the latent space, as well as an encoder that outputs a sparse set of linear coefficients for approximately reconstructing vectors in the latent space. One common SAE variant is TopK, which takes the top-K matching decoder features after a linear encoder projection \cite{gao2024scaling}.

\cite{bills2023language} found that LLMs can find explanations for MLP neurons in smaller networks and accurately predict (\textit{simulate}) their activations based on the explanations. SAE features can similarly be automatically explained and simulated \cite{huben2024sparse,bricken2023monosemanticity}, with scores exceeding those of MLP neuron features and other sparse decomposition techniques. \cite{eleuther_auto-2024} replaces the expensive per-token simulation step with classification of images into ones matching the explanation or not.

Denoising diffusion \cite{DBLP:journals/corr/abs-2006-11239,DBLP:conf/iclr/SongME21} is a generative modelling method that learns a score function of the original distribution mixed with varying amounts of noise. Flow matching and rectified flows \cite{lipman2023flowmatchinggenerativemodeling,liu2022flowstraightfastlearning} are similar but simpler formulations that learn the expected velocity of a particle in an SDE at any given timestep. Models trained in this formulation are what we will consider in this paper because they comprise most of the state-of-the-art text-to-image generators. Transformers have been adapted for image diffusion with adaptations to the architecture \cite{peebles2023scalablediffusionmodelstransformers}, including text-to-image generation \cite{esser2024scalingrectifiedflowtransformers} with the multimodal diffusion transformer (MMDiT) architecture. \flux \ is a recent MMDiT transformer with 12 billion parameters that achieves state-of-the-art performance. It has a step-distilled variant, \flux \ Schnell, that can generate images in one timestep.

SAEs have been applied to vision transformers previously, with activations similarly being taken from the residual stream: \cite{multimodalvit,steeringimages}. Some have been trained on class-conditioned and text-to-image diffusion models \cite{cywiński2025saeuroninterpretableconceptunlearning,kim2024textitreveliointerpretingleveragingsemantic,ijishakin2024hspace}. Some are trained on the innermost bottleneck of a diffusion U-Net \cite{ijishakin2024hspace} (see \cite{kwon2023diffusionmodelssemanticlatent} for evidence that this site is especially interpretable), others are spaced through the network; they are trained either on a single timestep or all of them distributed equally. A priori, we may expect that features activate specifically on some timesteps or spatial positions, and that different types of features are more or less frequent at earlier or later layers. With MMDiT, we can apply the same dictionary learning technique to different layers with similar results.

While we're not aware of any work applying autointerpretation to diffusion models, \cite{shaham2025multimodalautomatedinterpretabilityagent} is an example of advanced automated interpretability for vision models. The agent presented in the work has many functions, explanation and scoring being just a small subset of its capabilities.

\cite{turner_steering_2024} introduces the technique of adding vectors to the residual stream for manipulating transformer internals -- though there is a rich history 
of applying the same technique for GANs and diffusion models mentioned in the paper. There are many works on steering diffusion models with feature-like disentangled representations, discovered through training or unsupervised methods \cite{gandikota2023conceptslidersloraadaptors,park2023unsuperviseddiscoverysemanticlatent,kwon2023diffusionmodelssemanticlatent}.

Steering \textit{with SAE features} is a natural application of results from dictionary learning. \cite{chalnev_improving_2024,steeringimages,obrien_steering_2024} explore it in various contexts like steering language models for various behaviors or controlling CLIP-conditioned image generators with a global latent addition.

\cite{surkov2024unpackingsdxlturbointerpreting} trains SAEs on various hidden layers of the Stable Diffusion XL Turbo UNet \citep{sauer2023adversarialdiffusiondistillation,ronneberger2015unetconvolutionalnetworksbiomedical}. It thoroughly evaluates steering and even shows image generation from pure feature guidance. We scale up a similar approach, identify unique challenges arising for the MMDiT architecture of \flux (normalization) and show successful examples of steering. \cite{kim2025conceptsteerersleveragingksparse} trains SAEs on both SDXL-Turbo and \flux \ Dev, showing controllability comparable to or exceeding non-steering-based baselines in real-world scenarios under adversarial attacks. Unlike our work, they train SAEs on the text encoder of the model, thus affecting the conditioning instead of the generation process. This site for inserting SAEs is less likely to show concepts used by the vision transformer or enable circuits-based mechanistic interpretability \citep{cammarata2020thread:}.

Inference-time decomposition of activations (ITDA) \cite{itda} has been proposed as a modification to SAEs. It removes the learnable encoder and replaces it with ITO \cite{ITO}, also known as gradient pursuit \cite{4480155}. Instead of being trained with gradient descent, decoder rows are selected from training input batches according to a reconstruction loss criterion with an additional pruning step. ITDA promises significant efficiency gains, being able to train models in several minutes -- orders of magnitude faster than SAEs. It also offers more straightforward interpretability of latents, as each one corresponds to a token which may be traced to the SAE training dataset.






\section{Methodology}

We train SAEs and ITDAs on various layers and timesteps of the \flux\cite{flux2024} model with varying amounts of inference-time and training time resources granted to the SAEs/ITDA. In this section, we discuss the algorithms we use and the modifications we made to them for training. We give a full discussion of background research such as SAEs, diffusion and steering in the related work section (\Cref{sec:related}).

\begin{figure*}
    \centering
    \includegraphics[width=\linewidth]{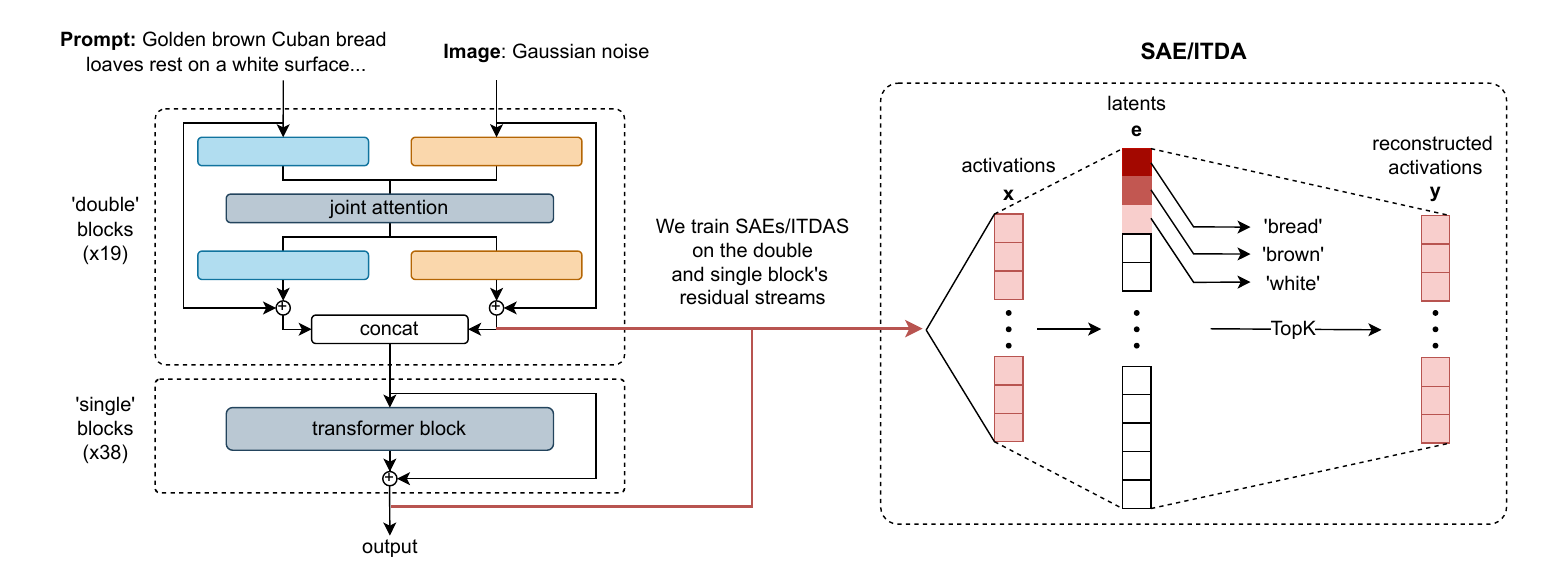}
    \caption{Residual SAEs for \flux}
    \label{fig:where-its-inserted}
\end{figure*}

\subsection{SAE formulation}

\label{sec:sae-formulation}

We build on TopK SAEs \cite{gao2024scaling} due to their simplicity. The TopK SAE encodes input vectors $x \in\mathbb{R}^n$ into sparse vectors $e \in \mathbb{R}^d: L_0(e) \le k$, where $k$ is the SAE's sparsity constant. The encoding process is usually affine. In our implementation, it is $e = \text{Top}_k(W_{enc} (x - b_{post} + b_{pre}))$, where $W_{enc} \in \mathbb{R}^{d \times n}$ and the biases $b_{post}, b_{pre} \in \mathbb{R}^n$. $k$ is a hyperparameter that controls the complexity of the problem, with higher $k$ allowing better reconstruction, but also potentially decreasing interpretability. The vector $e$ is then decoded into $y \in \mathbb{R}^d, y \approx x$, through an affine decoder: $y =W_{dec} e + b_{dec}$. This SAE is trained with MSE reconstruction loss against the original activations $\text{MSE}(x,y) = \sum_{i=1}^n \frac{1}{n} (y_i-x_i)^2$ with no additional penalties like AuxK.\footnote{\citet{gao2024scaling} trains with an auxiliary loss for avoiding ``dead features" -- encoder latent space dimensions that activate rarely. We also found AuxK helpful; see \Cref{sec:impl}.}

\subsection{ITDA formulation}

\citet{ITO}'s Inference-Time Optimization (\textsc{ITO}) is the algorithm behind ITDA. Originally, \textsc{ITO} was proposed as a way of replacing the SAE encoder -- finding sparse encoded vectors $e$ without learning the encoder. Using an efficient implementation of gradient pursuit \cite{4480155}, ITO can encode signals without encoder weights: $e = ITO(x, W_{dec})$. For further details, see \cite{ITO}.

ITDA builds on ITO by providing a simple algorithm for learning a dictionary without training an SAE. In short, it adds elements directly from the training dataset. It keeps the dictionary size low by prioritizing data points which are reconstructed poorly by the current version of the ITDA autoencoder. Taken together: $W_{dec,t+1} =W_{dec,t} \cup \left\{ x|x \in \text{data} \land \text{MSE}(x, ITO(x,W_{dec,t})) > \text{threshold} \right\}$, where $\text{threshold}$ is a hyperparameter we find using sweeps.

\subsection{SAE training}
\label{sec:how-sae}

For training SAEs and ITDAs, there are a few input parameters that determine the learning problem and its complexity. These include the input distribution -- determined by the site at which activations are taken, the resolution, the time step, as well as the distribution of input text -- and the SAE's capacity -- the $k$ and $d/n$ parameters.\footnote{Where $k$ is the sparsity from \Cref{sec:sae-formulation} and $d/n$ is expansion factor -- the dictionary size divided by the hidden size.}

We train most of our SAEs on single-step generations from \flux \ Schnell\footnote{One of the two released \flux models. Schnell is timestep-distilled and can generate images in 1 step. Dev is guidance-distilled and requires $\ge20$ timeseps.} at 256x256 resolution given resource constraints. We focus on layer 18 of the double blocks for many of the sweeps, as it is halfway through the model's parameter count. We also train on layers 9 (double and single) and layer 18 (single) (\Cref{fig:where-its-inserted}).
We train all SAEs for 30k steps, or 30M tokens.

The variance of \flux \ residual streams is largely concentrated in the first few dozen eigenvalues (\Cref{sec:basics:spectrum}). We found that standard training SAEs on residual stream data produces many dead features ($>99\%$), regardless of whether we use AuxK or not. While these autoencoders can have high proportions of variance explained ($>60\%$), the features will be concentrated in one low-dimensional subspace, similarly to the training data.

We propose a solution: normalizing the spectral components of the inputs before the SAE through PCA whitening in input space (see \Cref{app:pca-norm}). We compare outputs for training in the whitened space, but for inference we transform reconstructions back to the original residual stream: $y^* = W^T (y \odot \vec{\sigma} (x_1) + \vec{\mu} (x_2))$.\footnote{This is an affine transformation that can be ``folded into" the weights of the SAE (as in \citet{rajamanoharan2024jumpingaheadimprovingreconstruction}, Appendix A). }

Training with this normalization produces manageable amounts of dead latents ($<30\%$). In other experiments, we found that normalizing by the mean and standard deviation without PCA has similar performance, but all our SAE work here also does the PCA transformation.

We repeat that normalization is necessary for training Flux SAEs without the majority of neurons being dead due high anisotropy of the residual stream (\Cref{sec:basics:spectrum}).

\subsection{PCA normalization}
\label{app:pca-norm}
Based on the first few batches of the training data, we compute the orthonormal projection matrix of the PCA decomposition $W \in O(n)$. We use it to transform the data ($\vec{x_1} = W \vec{x_0}$) into a space where the components have zero covariance, anisotropic variance, and an unknown, potentially large mean. We further remove the mean and variance through standardization ($\vec{x_2} = \vec{x_1} \odot \frac{1}{\vec{\sigma} (x_1)}$, $\vec{x_3} = \vec{x_2} - \vec{\mu} (x_2)$). In sum, we perform whitening of the input data.

\subsection{ITDA training}

We similarly normalize training inputs to ITDA without PCA. We tried several modifications to ITDA: restricting feature growth early on by taking the top-k feature additions; adding the residual reconstruction error instead of the datapoint to the dictionary; pruning the ITDA after training with K-Means clustering. None of them significantly improved on the dictionary size to reconstruction accuracy tradeoff.

We use the FVU (fraction of variance unexplained) metric as the loss function: $\text{FVU}(y,y^*) = \text{MSE}(y, y^*) / \text{MSE}(y^*, \mu_{y^*})$, where $y^*$ is the ground truth. This metric can go above 1, but typically takes values from 0.1 to 0.5.

\section{Basic Flux interpretability}
\label{sec:basics}


While \flux \ uses the Transformer architecture,
there are differences that could cause techniques applicable to Transformers to not work with it. For example: Flux uses AdaLN layers \cite{huang2017arbitrarystyletransferrealtime}, and Layernorm layers are commonly known to hamper interpretability because they are nonlinear and can cause anomalies like \cite{kovaleva2021bertbustersoutlierdimensions}; a more complex version of LN could cause more issues because of the timestep and prompt dependence. Similarly, the gradients induced by the architecture and data distribution can render some optimizers less effective and potentially cause artifacts \cite{zhang2020adaptivemethodsgoodattention}.

\subsection{Across layers and timesteps}

\begin{figure}
    \centering
    \includegraphics[width=\linewidth]{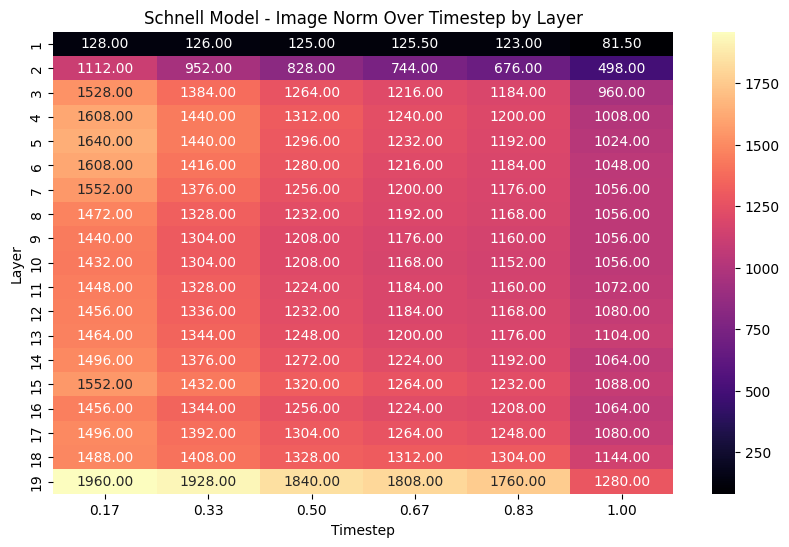}
    \caption{Residual stream norms for double blocks}
    \label{fig:norm-double}
\end{figure}

\begin{figure}
    \centering
    \includegraphics[width=\linewidth]{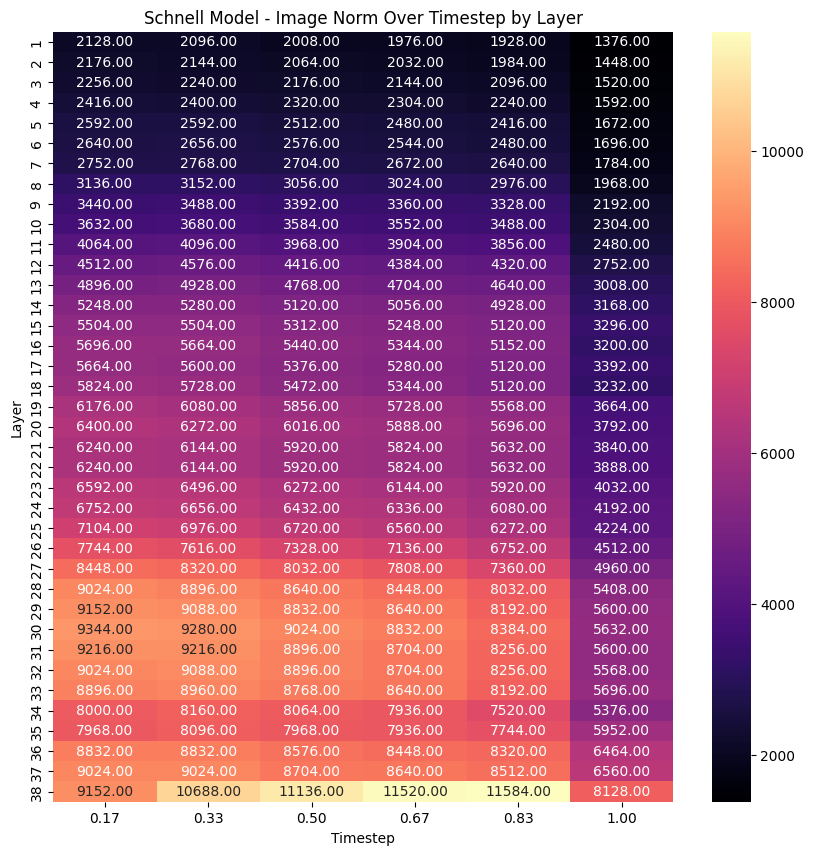}
    \caption{Residual stream norms for single blocks}
    \label{fig:norm-single}
\end{figure}

We start by comparing the norms of the residual stream throughout the layers. At the last layer of the double blocks (\Cref{fig:norm-double}), the norm of the text component jumps up rapidly. This may correspond to a distinct stage of inference \cite{lad2024remarkablerobustnessllmsstages}.

The residual stream norms generally increase through timesteps, but they only rise by an order of magnitude.

\subsection{Latent space spectrum}
\label{sec:basics:spectrum}

When we trained SAEs without any normalization (\Cref{sec:how-sae}), we found adequate reconstruction quality with many dead features. The alive features seemed interpretable, but there were less than 1000 of them. This suggests that there may be a small subspace in the residual stream that contains most of the variance for the SAE to explain.

\begin{figure}
    \centering
    \includegraphics[width=\linewidth]{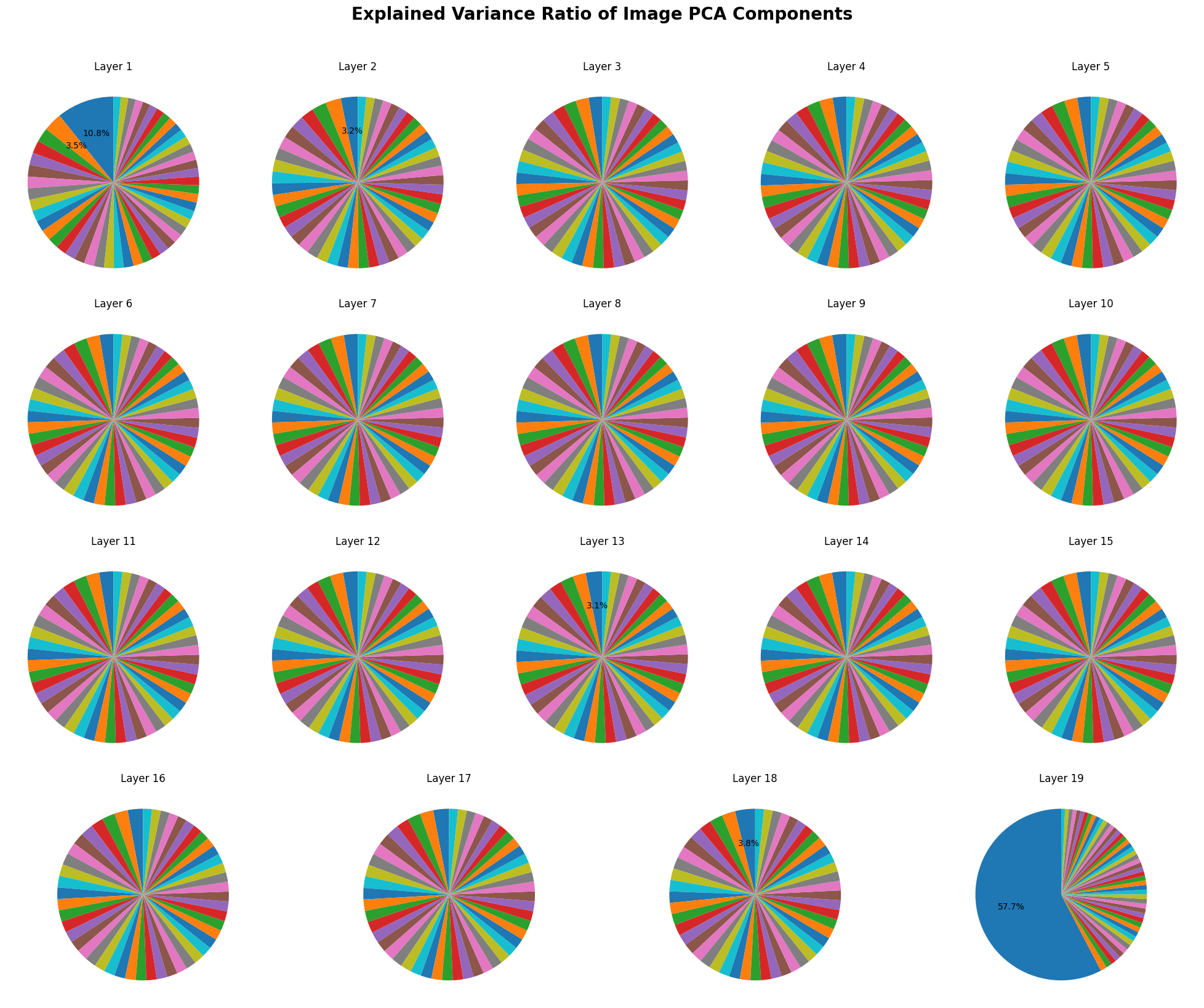}
    \caption{Variances explained by principal components}
    \label{fig:principal-components}
\end{figure}

We measure the variances of each PCA component (\Cref{fig:principal-components}) and residual stream element, and find that they are \textit{anisotropic} to an extent that would be unusual for a text model. That these dimensions containing most of the variance are basis-aligned is perplexing, and may be related to outlier dimensions \cite{kovaleva2021bertbustersoutlierdimensions}. It's possible they encode CLIP image embeddings, the model's encoding for noise in the VAE space,\footnote{This explains the low rank but not the interpretability of these latents.} or positional embeddings.

\section{Evaluations}
\label{sec:evals}

\subsection{Reconstruction}

\begin{figure}
    \centering
    \includegraphics[trim=0cm 0cm 1.2cm 0cm,clip=true,width=\linewidth]{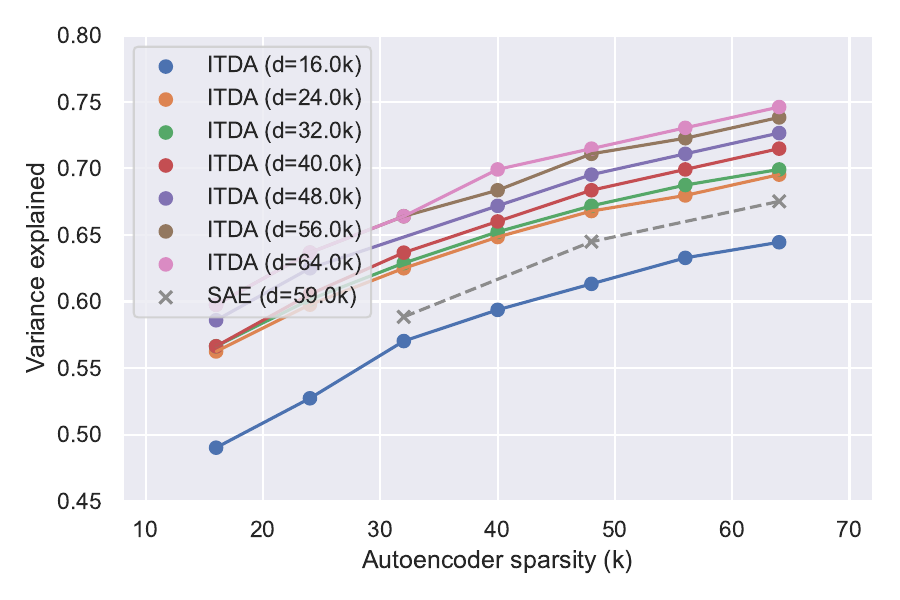}
    \caption{Reconstruction quality of \flux \ double block 18 residual SAEs and ITDAs.}
    \label{fig:l18-fvu}
\end{figure}
The primary objective of SAEs is reconstructing activations. We evaluate ITDAs and SAEs on using the \textbf{variance explained} metric: $\text{VE}(y,y^*) = \sum_{i=0}^n \frac{1}{\sigma_y^2 \sigma_{y^*}^2} \left( \frac{1}{|y|} \sum_{j=0}^{|y|} (y - \mu_y)_{j,i} (y^* - \mu_{y^*})_{j,i} \right)^2$.

As mentioned above, we compare SAEs and ITDAs in two settings: comparing various settings of $k$ and $d/n$ on layer 18, and testing the performance on various layers in the network. \Cref{fig:l18-fvu} demonstrates the former, showing that ITDA is generally superior in terms of reconstruction performance at similar dictionary sizes.

We compare FVU scores across different layers in \Cref{more-stats}.

\subsection{Automated interpretability}

\citet{bills2023language} introduced automated interpretability for feedforward block (MLP, \cite{vaswani2023attentionneed}) neurons: a pipeline that uses a language model to generate explanations for maximum activating examples of neurons, and then predicts the strength of the neuron's activations on a test set. Detection scoring \cite{eleuther_auto-2024} replaces the regression task of predicting how strongly a neuron activates on each token with the classification task of detecting whether the neuron was correctly labeled, reducing the token usage of the autointerpretation pipeline.

Since we are not working with raw language models, unlike this prior work, we need to adapt these autointerpretability methods. We introduce a visual autointerpretation pipeline with two components: the explainer and the classifier. The explainer looks at images with activations painted in blue on top of them and generates a shared explanation. The scorer looks at a single image and decides if it belongs to the chosen explanation. Both the explainer and the scorer use google/gemini-2.0-flash-001 as the backend. Their prompts are included in \Cref{sec:autointerp-prompt}.

We run autointerp on layer 18, $k=64$, $d=64000$. We compare SAEs and ITDAs to an MLP neuron baseline, similarly to \cite{huben2024sparse}. We use max-activating examples and only consider activating features from each method. We see that SAEs and ITDA have comparable performance and that MLP neurons are generally less interpretable, although they contain several features with exceedingly high autointerp scores. From visual inspection (\Cref{more-activations}), highly interpretable MLP neurons activate on contiguous areas of the image with fixed semantic meaning and are likely causally relevant to the image generation process.

\begin{figure}
    \centering
    \includegraphics[width=\linewidth]{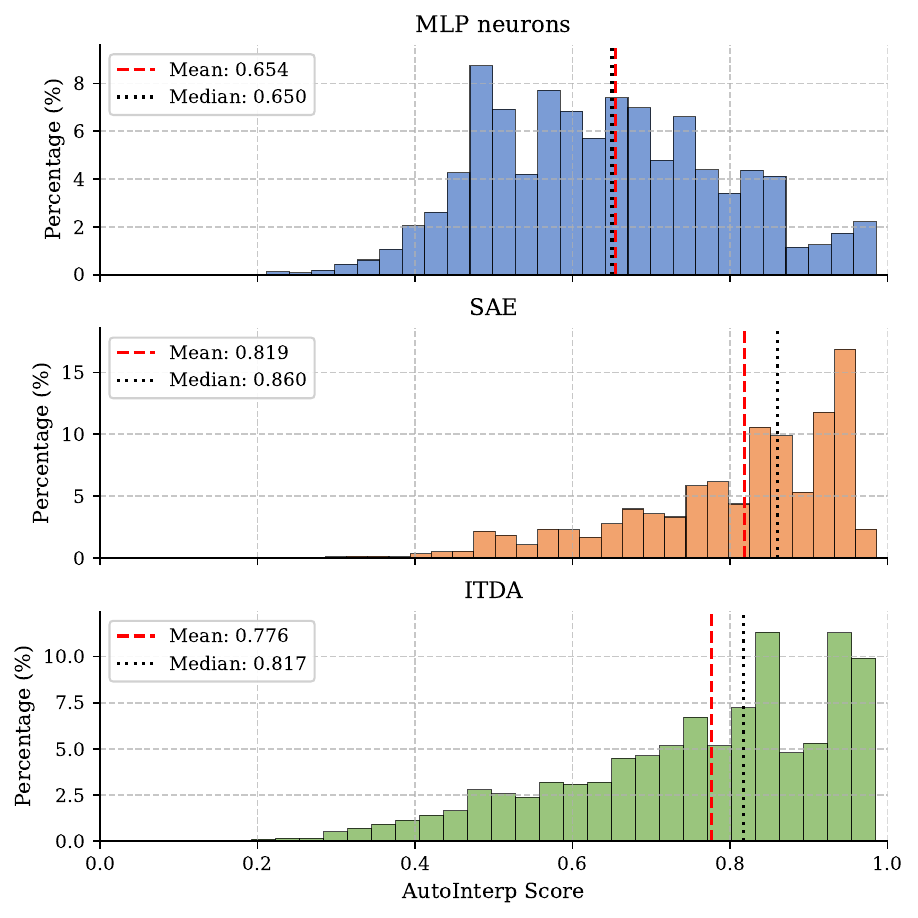}
    \caption{Histograms of autointerp accuracy scores for three methods.}
    \label{fig:autointerp}
\end{figure}



\subsection{Steering}

As mentioned in \Cref{sec:related}, steering has a rich history in diffusion models, and steering with SAE features is a common technique. We explore steering \flux \ with our trained SAEs at the intended layers through simple activation addition at areas of the image: $y[a \mathord{:} b,c \mathord{:}d] \mathrel{+}= W_{dec}[f]$.

We primarily study Layer 18. We found that steering with features often leads to an effect associated with their max activating patterns. However, the effects are constrained, in a sense that the initial prompt should be related to the feature: for example the anime style feature from \Cref{fig:activations} had effect only on the "A cartoon" prompt (but not on a prompt like "A person"). This suggests that images may have a localized feature space, and steering may work only inside them.

We steered only on a fraction of the steps. In the examples (\Cref{fig:activations}), we applied steering only to the first 5 steps out of 7. Steering on all steps introduced noticeable artifacts. Negative steering was also successful in some cases, although it could remove only small details of the image and required much more complex prompts than positive steering.
\section{Future Work}
\label{sec:future}


We evaluated our SAEs and ITDAs on simple metrics like autointerp. In vision, there are many tasks, like classification, segmentation and depth estimation, that SAEs could be used to perform \citep{chen2023surfacestatisticsscenerepresentations}; this would be analogous to sparse probing \cite{gao2024scaling}.

In this work, we apply the SAE architecture as used in large language models (LLMs) without modifications. However, it is reasonable to expect some changes to be necessary: in images, latent activations of nearby spatial positions are more similar than those of random patches.\footnote{Especially if they correspond to semantically meaningful contiguous regions of the image. See \cite{nostalgebraist} for discussion of a similar problem with LLMs.} We could adapt the encoder architecture to add spatial inductive bias, like by making it convolutional. We could improve steering by making the process more similar to finetuning \cite{kwon2023diffusionmodelssemanticlatent,zhang2023addingconditionalcontroltexttoimage} or otherwise aware of the effects of the steering.

We only consider SAEs on image activations in this work. It is likely that the multimodal DiT architecture shares features between text and image streams, which is something that could be the cosine similarity of SAEs. This paper also does not consider variable image resolution and timesteps other than pure noise. \citet{carter2019activation}'s Activation Atlas could be a useful technique for dealing with intermediate timesteps.

Another promising avenue for future work is comparing \flux's two variants, Dev and Schnell. We only considered the latter, but crosscoders \cite{lindsey2024sparse} may let us find corresponding pairs of features and features unique to step-distilled models.

It is also important to investigate biases specific to our autointerp pipeline. Similarly to \citet{heap2025sparseautoencodersinterpretrandomly}, it is possible that our pipeline may pick up on simple characteristics of the image like color. We leave detailed investigation to future work.

\section{Conclusion}
\label{sec:conclusion}

In this work, we have demonstrated the successful application of Sparse Autoencoders (SAEs) and Inference-Time Decomposition of Activations (ITDA) to large text-to-image diffusion models. Our experiments with \flux \ show that these methods can effectively decompose complex residual stream activations into interpretable features that enable targeted steering of image generation. We find that both SAEs and ITDAs outperform MLP neurons on interpretability metrics, while maintaining high reconstruction quality across various model layers and configurations.

Our manual examination revealed important distinctions between SAEs and ITDAs not captured by automated metrics. Despite similar autointerp scores, ITDA features typically encoded general attributes (colors, textures, regions) while SAE features captured more concrete objects (hats, faces). This qualitative difference highlights a limitation of autointerp metrics - they don't distinguish between abstract visual properties and manipulable semantic concepts. SAE features also demonstrated superior pixel-wise coverage compared to the sparser activation patterns of ITDA features (\Cref{more-activations}).

When exploring steering capabilities, we observed that SAE feature steering was effective but constrained, requiring initial prompts to align with steered features. This suggests a localized nature of feature representation in diffusion models, where later layer features have low effect if they are not directly related to the image representation generated by earlier layers.
{
    \small
    \bibliographystyle{ieeenat_fullname}
    \bibliography{main}

\begin{thebibliography}{53}
\providecommand{\natexlab}[1]{#1}
\providecommand{\url}[1]{\texttt{#1}}
\expandafter\ifx\csname urlstyle\endcsname\relax
  \providecommand{\doi}[1]{doi: #1}\else
  \providecommand{\doi}{doi: \begingroup \urlstyle{rm}\Url}\fi

\bibitem[Bills et~al.(2023)Bills, Cammarata, Mossing, Tillman, Gao, Goh, Sutskever, Leike, Wu, and Saunders]{bills2023language}
Steven Bills, Nick Cammarata, Dan Mossing, Henk Tillman, Leo Gao, Gabriel Goh, Ilya Sutskever, Jan Leike, Jeff Wu, and William Saunders.
\newblock Language models can explain neurons in language models.
\newblock \emph{URL https://openaipublic. blob. core. windows. net/neuron-explainer/paper/index. html.(Date accessed: 14.05. 2023)}, 2, 2023.

\bibitem[Blumensath and Davies(2008)]{4480155}
Thomas Blumensath and Mike~E. Davies.
\newblock Gradient pursuits.
\newblock \emph{IEEE Transactions on Signal Processing}, 56\penalty0 (6):\penalty0 2370--2382, 2008.

\bibitem[Bricken et~al.(2023)Bricken, Templeton, Batson, Chen, Jermyn, Conerly, Turner, Anil, Denison, Askell, Lasenby, Wu, Kravec, Schiefer, Maxwell, Joseph, Hatfield-Dodds, Tamkin, Nguyen, McLean, Burke, Hume, Carter, Henighan, and Olah]{bricken2023monosemanticity}
Trenton Bricken, Adly Templeton, Joshua Batson, Brian Chen, Adam Jermyn, Tom Conerly, Nick Turner, Cem Anil, Carson Denison, Amanda Askell, Robert Lasenby, Yifan Wu, Shauna Kravec, Nicholas Schiefer, Tim Maxwell, Nicholas Joseph, Zac Hatfield-Dodds, Alex Tamkin, Karina Nguyen, Brayden McLean, Josiah~E Burke, Tristan Hume, Shan Carter, Tom Henighan, and Christopher Olah.
\newblock Towards monosemanticity: Decomposing language models with dictionary learning.
\newblock \emph{Transformer Circuits Thread}, 2023.
\newblock https://transformer-circuits.pub/2023/monosemantic-features/index.html.

\bibitem[Cammarata et~al.(2020)Cammarata, Carter, Goh, Olah, Petrov, Schubert, Voss, Egan, and Lim]{cammarata2020thread:}
Nick Cammarata, Shan Carter, Gabriel Goh, Chris Olah, Michael Petrov, Ludwig Schubert, Chelsea Voss, Ben Egan, and Swee~Kiat Lim.
\newblock Thread: Circuits.
\newblock \emph{Distill}, 2020.
\newblock https://distill.pub/2020/circuits.

\bibitem[Carlini et~al.(2023)Carlini, Hayes, Nasr, Jagielski, Sehwag, Tramèr, Balle, Ippolito, and Wallace]{carlini2023extracting}
Nicholas Carlini, Jamie Hayes, Milad Nasr, Matthew Jagielski, Vikash Sehwag, Florian Tramèr, Borja Balle, Daphne Ippolito, and Eric Wallace.
\newblock Extracting {Training} {Data} from {Diffusion} {Models}.
\newblock \emph{UseNix Security}, 2023.
\newblock arXiv:2301.13188 [cs].

\bibitem[Carter et~al.(2019)Carter, Armstrong, Schubert, Johnson, and Olah]{carter2019activation}
Shan Carter, Zan Armstrong, Ludwig Schubert, Ian Johnson, and Chris Olah.
\newblock Activation atlas.
\newblock \emph{Distill}, 2019.
\newblock https://distill.pub/2019/activation-atlas.

\bibitem[Chalnev et~al.(2024)Chalnev, Siu, and Conmy]{chalnev_improving_2024}
Sviatoslav Chalnev, Matthew Siu, and Arthur Conmy.
\newblock Improving {Steering} {Vectors} by {Targeting} {Sparse} {Autoencoder} {Features}, 2024.
\newblock arXiv:2411.02193 [cs].

\bibitem[Chen et~al.(2023)Chen, Viégas, and Wattenberg]{chen2023surfacestatisticsscenerepresentations}
Yida Chen, Fernanda Viégas, and Martin Wattenberg.
\newblock Beyond surface statistics: Scene representations in a latent diffusion model, 2023.

\bibitem[Cunningham et~al.(2024)Cunningham, Ewart, Riggs, Huben, and Sharkey]{cunningham}
Hoagy Cunningham, Aidan Ewart, Logan Riggs, Robert Huben, and Lee Sharkey.
\newblock Sparse autoencoders find highly interpretable features in language models.
\newblock \emph{arXiv preprint arXiv:2401.01345}, 2024.

\bibitem[Cywiński and Deja(2025)]{cywiński2025saeuroninterpretableconceptunlearning}
Bartosz Cywiński and Kamil Deja.
\newblock Saeuron: Interpretable concept unlearning in diffusion models with sparse autoencoders, 2025.

\bibitem[Daujotas(2024)]{steeringimages}
Gytis Daujotas.
\newblock Interpreting and steering features in images.
\newblock \emph{LessWrong}, 2024.

\bibitem[Dettmers et~al.(2023)Dettmers, Pagnoni, Holtzman, and Zettlemoyer]{dettmers2023qloraefficientfinetuningquantized}
Tim Dettmers, Artidoro Pagnoni, Ari Holtzman, and Luke Zettlemoyer.
\newblock Qlora: Efficient finetuning of quantized llms, 2023.

\bibitem[Elhage et~al.(2022)Elhage, Hume, Olsson, Schiefer, Henighan, Kravec, Hatfield-Dodds, Lasenby, Drain, Chen, Grosse, McCandlish, Kaplan, Amodei, Wattenberg, and Olah]{elhage2022superposition}
Nelson Elhage, Tristan Hume, Catherine Olsson, Nicholas Schiefer, Tom Henighan, Shauna Kravec, Zac Hatfield-Dodds, Robert Lasenby, Dawn Drain, Carol Chen, Roger Grosse, Sam McCandlish, Jared Kaplan, Dario Amodei, Martin Wattenberg, and Christopher Olah.
\newblock Toy models of superposition.
\newblock \emph{Transformer Circuits Thread}, 2022.
\newblock https://transformer-circuits.pub/2022/toy\_model/index.html.

\bibitem[Esser et~al.(2024)Esser, Kulal, Blattmann, Entezari, Müller, Saini, Levi, Lorenz, Sauer, Boesel, Podell, Dockhorn, English, Lacey, Goodwin, Marek, and Rombach]{esser2024scalingrectifiedflowtransformers}
Patrick Esser, Sumith Kulal, Andreas Blattmann, Rahim Entezari, Jonas Müller, Harry Saini, Yam Levi, Dominik Lorenz, Axel Sauer, Frederic Boesel, Dustin Podell, Tim Dockhorn, Zion English, Kyle Lacey, Alex Goodwin, Yannik Marek, and Robin Rombach.
\newblock Scaling rectified flow transformers for high-resolution image synthesis, 2024.

\bibitem[Fry(2024)]{multimodalvit}
Hugo Fry.
\newblock Towards multimodal interpretability: Learning sparse interpretable features in vision transformers.
\newblock \emph{LessWrong}, 2024.

\bibitem[Gandikota et~al.(2023)Gandikota, Materzynska, Zhou, Torralba, and Bau]{gandikota2023conceptslidersloraadaptors}
Rohit Gandikota, Joanna Materzynska, Tingrui Zhou, Antonio Torralba, and David Bau.
\newblock Concept sliders: Lora adaptors for precise control in diffusion models, 2023.

\bibitem[Gao et~al.(2024)Gao, la~Tour, Tillman, Goh, Troll, Radford, Sutskever, Leike, and Wu]{gao2024scaling}
Leo Gao, Tom~Dupr{\'e} la Tour, Henk Tillman, Gabriel Goh, Rajan Troll, Alec Radford, Ilya Sutskever, Jan Leike, and Jeffrey Wu.
\newblock Scaling and evaluating sparse autoencoders.
\newblock \emph{arXiv preprint arXiv:2406.04093}, 2024.

\bibitem[Heap et~al.(2025)Heap, Lawson, Farnik, and Aitchison]{heap2025sparseautoencodersinterpretrandomly}
Thomas Heap, Tim Lawson, Lucy Farnik, and Laurence Aitchison.
\newblock Sparse autoencoders can interpret randomly initialized transformers, 2025.

\bibitem[Ho et~al.(2020)Ho, Jain, and Abbeel]{DBLP:journals/corr/abs-2006-11239}
Jonathan Ho, Ajay Jain, and Pieter Abbeel.
\newblock Denoising diffusion probabilistic models.
\newblock \emph{CoRR}, abs/2006.11239, 2020.

\bibitem[Huang and Belongie(2017)]{huang2017arbitrarystyletransferrealtime}
Xun Huang and Serge Belongie.
\newblock Arbitrary style transfer in real-time with adaptive instance normalization, 2017.

\bibitem[Huben et~al.(2024)Huben, Cunningham, Smith, Ewart, and Sharkey]{huben2024sparse}
Robert Huben, Hoagy Cunningham, Logan~Riggs Smith, Aidan Ewart, and Lee Sharkey.
\newblock Sparse autoencoders find highly interpretable features in language models.
\newblock In \emph{The Twelfth International Conference on Learning Representations}, 2024.

\bibitem[Ijishakin et~al.(2024)Ijishakin, Ang, Baljer, Tan, Fry, Abdulaal, Lynch, and Cole]{ijishakin2024hspace}
Ayodeji Ijishakin, Ming~Liang Ang, Levente Baljer, Daniel Chee~Hian Tan, Hugo~Laurence Fry, Ahmed Abdulaal, Aengus Lynch, and James~H. Cole.
\newblock H-space sparse autoencoders.
\newblock In \emph{Neurips Safe Generative AI Workshop 2024}, 2024.

\bibitem[Juang et~al.(2024)Juang, Paulo, Drori, and Nora]{eleuther_auto-2024}
Caden Juang, Gon\c{c}alo Paulo, Jacob Drori, and Belrosem Nora.
\newblock {Understanding and steering Llama 3}, 2024.

\bibitem[Kim and Ghadiyaram(2025)]{kim2025conceptsteerersleveragingksparse}
Dahye Kim and Deepti Ghadiyaram.
\newblock Concept steerers: Leveraging k-sparse autoencoders for controllable generations, 2025.

\bibitem[Kim et~al.(2024)Kim, Thomas, and Ghadiyaram]{kim2024textitreveliointerpretingleveragingsemantic}
Dahye Kim, Xavier Thomas, and Deepti Ghadiyaram.
\newblock $\textit{Revelio}$: Interpreting and leveraging semantic information in diffusion models, 2024.

\bibitem[Kovaleva et~al.(2021)Kovaleva, Kulshreshtha, Rogers, and Rumshisky]{kovaleva2021bertbustersoutlierdimensions}
Olga Kovaleva, Saurabh Kulshreshtha, Anna Rogers, and Anna Rumshisky.
\newblock Bert busters: Outlier dimensions that disrupt transformers, 2021.

\bibitem[Kramár(2024)]{kramar}
János Kramár.
\newblock Instrumenting llm model internals in jax, 2024.

\bibitem[Kwon et~al.(2023)Kwon, Jeong, and Uh]{kwon2023diffusionmodelssemanticlatent}
Mingi Kwon, Jaeseok Jeong, and Youngjung Uh.
\newblock Diffusion models already have a semantic latent space, 2023.

\bibitem[Labs(2024)]{flux2024}
Black~Forest Labs.
\newblock Flux.
\newblock \url{https://github.com/black-forest-labs/flux}, 2024.

\bibitem[Lad et~al.(2024)Lad, Gurnee, and Tegmark]{lad2024remarkablerobustnessllmsstages}
Vedang Lad, Wes Gurnee, and Max Tegmark.
\newblock The remarkable robustness of llms: Stages of inference?, 2024.

\bibitem[Leask et~al.(2025)Leask, Nanda, and Moubayed]{itda}
Patrick Leask, Neel Nanda, and Noura~Al Moubayed.
\newblock Inference-{Time} {Decomposition} of {Activations} ({ITDA}): {A} {Scalable} {Approach} to {Interpreting} {Large} {Language} {Models}.
\newblock \emph{ICML 2025}, 2025.
\newblock arXiv:2505.17769 [cs] version: 1.

\bibitem[Lindsey* et~al.(2024)Lindsey*, Templeton*, Marcus*, Conerly*, Batson, and Olah]{lindsey2024sparse}
Jack Lindsey*, Adly Templeton*, Jonathan Marcus*, Thomas Conerly*, Joshua Batson, and Christopher Olah.
\newblock Sparse crosscoders for cross-layer features and model diffing.
\newblock \emph{Transformer Circuits}, 2024.

\bibitem[Lipman et~al.(2023)Lipman, Chen, Ben-Hamu, Nickel, and Le]{lipman2023flowmatchinggenerativemodeling}
Yaron Lipman, Ricky T.~Q. Chen, Heli Ben-Hamu, Maximilian Nickel, and Matt Le.
\newblock Flow matching for generative modeling, 2023.

\bibitem[Liu et~al.(2022)Liu, Gong, and Liu]{liu2022flowstraightfastlearning}
Xingchao Liu, Chengyue Gong, and Qiang Liu.
\newblock Flow straight and fast: Learning to generate and transfer data with rectified flow, 2022.

\bibitem[Ng(2011)]{andrewng}
Andrew Ng.
\newblock Sparse autoencoder.
\newblock \emph{CS294A Lecture Notes}, 2011.
\newblock Unpublished lecture notes.

\bibitem[nostalgebraist(2024)]{nostalgebraist}
nostalgebraist.
\newblock Shortform comment on sae locality.
\newblock \emph{LessWrong}, 2024.

\bibitem[O'Brien et~al.(2024)O'Brien, Majercak, Fernandes, Edgar, Chen, Nori, Carignan, Horvitz, and Poursabzi-Sangde]{obrien_steering_2024}
Kyle O'Brien, David Majercak, Xavier Fernandes, Richard Edgar, Jingya Chen, Harsha Nori, Dean Carignan, Eric Horvitz, and Forough Poursabzi-Sangde.
\newblock Steering {Language} {Model} {Refusal} with {Sparse} {Autoencoders}.
\newblock 2024.
\newblock Publisher: arXiv Version Number: 1.

\bibitem[Park et~al.(2023)Park, Kwon, Jo, and Uh]{park2023unsuperviseddiscoverysemanticlatent}
Yong-Hyun Park, Mingi Kwon, Junghyo Jo, and Youngjung Uh.
\newblock Unsupervised discovery of semantic latent directions in diffusion models, 2023.

\bibitem[Peebles and Xie(2023)]{peebles2023scalablediffusionmodelstransformers}
William Peebles and Saining Xie.
\newblock Scalable diffusion models with transformers, 2023.

\bibitem[Rajamanoharan et~al.(2024)Rajamanoharan, Lieberum, Sonnerat, Conmy, Varma, Kramár, and Nanda]{rajamanoharan2024jumpingaheadimprovingreconstruction}
Senthooran Rajamanoharan, Tom Lieberum, Nicolas Sonnerat, Arthur Conmy, Vikrant Varma, János Kramár, and Neel Nanda.
\newblock Jumping ahead: Improving reconstruction fidelity with jumprelu sparse autoencoders, 2024.

\bibitem[Ronneberger et~al.(2015)Ronneberger, Fischer, and Brox]{ronneberger2015unetconvolutionalnetworksbiomedical}
Olaf Ronneberger, Philipp Fischer, and Thomas Brox.
\newblock U-net: Convolutional networks for biomedical image segmentation, 2015.

\bibitem[Sauer et~al.(2023)Sauer, Lorenz, Blattmann, and Rombach]{sauer2023adversarialdiffusiondistillation}
Axel Sauer, Dominik Lorenz, Andreas Blattmann, and Robin Rombach.
\newblock Adversarial diffusion distillation, 2023.

\bibitem[Shaham et~al.(2025)Shaham, Schwettmann, Wang, Rajaram, Hernandez, Andreas, and Torralba]{shaham2025multimodalautomatedinterpretabilityagent}
Tamar~Rott Shaham, Sarah Schwettmann, Franklin Wang, Achyuta Rajaram, Evan Hernandez, Jacob Andreas, and Antonio Torralba.
\newblock A multimodal automated interpretability agent, 2025.

\bibitem[Shan et~al.(2024)Shan, Ding, Passananti, Wu, Zheng, and Zhao]{shan2024nightshade}
Shawn Shan, Wenxin Ding, Josephine Passananti, Stanley Wu, Haitao Zheng, and Ben~Y. Zhao.
\newblock Nightshade: {Prompt}-{Specific} {Poisoning} {Attacks} on {Text}-to-{Image} {Generative} {Models}.
\newblock 2024.
\newblock arXiv:2310.13828 [cs].

\bibitem[Sharkey et~al.(2022)Sharkey, Braun, and Millidge]{takingfeatures}
Lee Sharkey, Dan Braun, and Beren Millidge.
\newblock Taking features out of superposition with sparse autoencoders.
\newblock \emph{AI Alignment Forum}, 2022.

\bibitem[Smith(2024)]{ITO}
Lewis Smith.
\newblock Replacing sae encoders with inference-time optimisation, 2024.

\bibitem[Song et~al.(2021)Song, Meng, and Ermon]{DBLP:conf/iclr/SongME21}
Jiaming Song, Chenlin Meng, and Stefano Ermon.
\newblock Denoising diffusion implicit models.
\newblock In \emph{9th International Conference on Learning Representations, {ICLR} 2021, Virtual Event, Austria, May 3-7, 2021}. OpenReview.net, 2021.

\bibitem[Surkov et~al.(2024)Surkov, Wendler, Terekhov, Deschenaux, West, and Gulcehre]{surkov2024unpackingsdxlturbointerpreting}
Viacheslav Surkov, Chris Wendler, Mikhail Terekhov, Justin Deschenaux, Robert West, and Caglar Gulcehre.
\newblock Unpacking sdxl turbo: Interpreting text-to-image models with sparse autoencoders, 2024.

\bibitem[Turner et~al.(2024)Turner, Thiergart, Leech, Udell, Vazquez, Mini, and MacDiarmid]{turner_steering_2024}
Alexander~Matt Turner, Lisa Thiergart, Gavin Leech, David Udell, Juan~J. Vazquez, Ulisse Mini, and Monte MacDiarmid.
\newblock Steering {Language} {Models} {With} {Activation} {Engineering}, 2024.
\newblock arXiv:2308.10248 [cs].

\bibitem[Vaswani et~al.(2023)Vaswani, Shazeer, Parmar, Uszkoreit, Jones, Gomez, Kaiser, and Polosukhin]{vaswani2023attentionneed}
Ashish Vaswani, Noam Shazeer, Niki Parmar, Jakob Uszkoreit, Llion Jones, Aidan~N. Gomez, Lukasz Kaiser, and Illia Polosukhin.
\newblock Attention is all you need, 2023.

\bibitem[Zhan et~al.(2024)Zhan, Zheng, Xie, and Zisserman]{zhan2024generalprotocolprobelarge}
Guanqi Zhan, Chuanxia Zheng, Weidi Xie, and Andrew Zisserman.
\newblock A general protocol to probe large vision models for 3d physical understanding, 2024.

\bibitem[Zhang et~al.(2020)Zhang, Karimireddy, Veit, Kim, Reddi, Kumar, and Sra]{zhang2020adaptivemethodsgoodattention}
Jingzhao Zhang, Sai~Praneeth Karimireddy, Andreas Veit, Seungyeon Kim, Sashank~J Reddi, Sanjiv Kumar, and Suvrit Sra.
\newblock Why are adaptive methods good for attention models?, 2020.

\bibitem[Zhang et~al.(2023)Zhang, Rao, and Agrawala]{zhang2023addingconditionalcontroltexttoimage}
Lvmin Zhang, Anyi Rao, and Maneesh Agrawala.
\newblock Adding conditional control to text-to-image diffusion models, 2023.

\end{thebibliography}
}

\clearpage
\setcounter{page}{1}
\maketitlesupplementary

\appendix

\newtcolorbox{modelprompt}[1][]{
  colback=gray!5,           
  colframe=gray!50!black,   
  boxrule=0.5pt,            
  sharp corners,            
  fonttitle=\bfseries,      
  title={Model Prompt},
  #1,
  breakable,                
  left=6pt,                 
  right=6pt,                
  top=6pt,                  
  bottom=6pt,               
  before skip=8pt,          
  after skip=8pt            
}



\section{Detailed evaluation statistics}
\begin{figure}
    \centering
    \includegraphics[width=\linewidth]{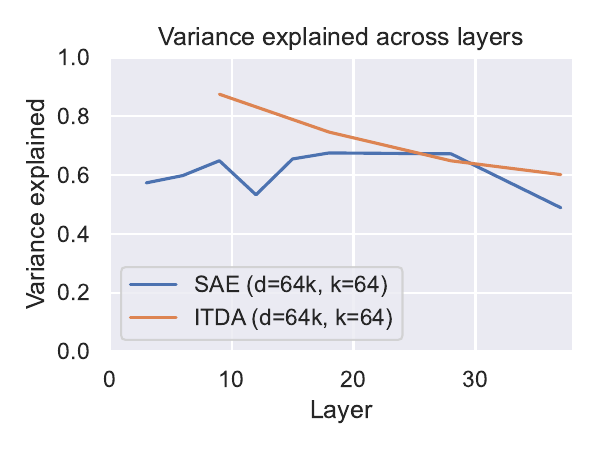}
    \caption{Reconstruction quality for ITDAs and SAEs across layers.}
    \label{fig:across-layers}
\end{figure}

In \Cref{fig:across-layers}, we compare ITDAs and SAEs across multiple layers.

\label{more-stats}
\section{Additional example feature activations}
\label{more-activations}

\begin{figure}
    \centering
    \includegraphics[width=\linewidth]{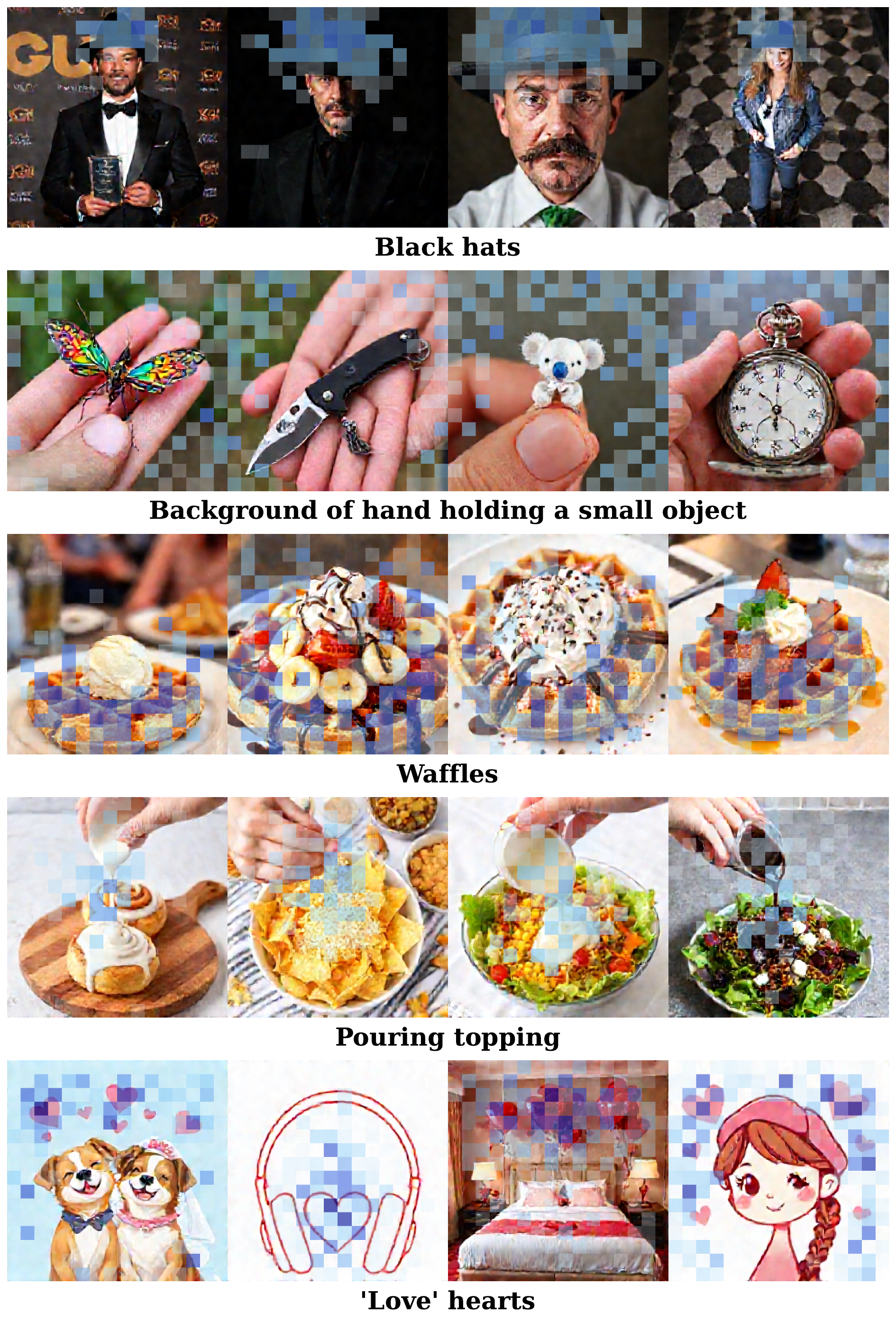}
    \caption{Maximum activating examples of some interpretable features from our \flux \ SAE. }
    \label{fig:maxacts-more-saes}
\end{figure}

\begin{figure}
    \centering
    \includegraphics[width=\linewidth]{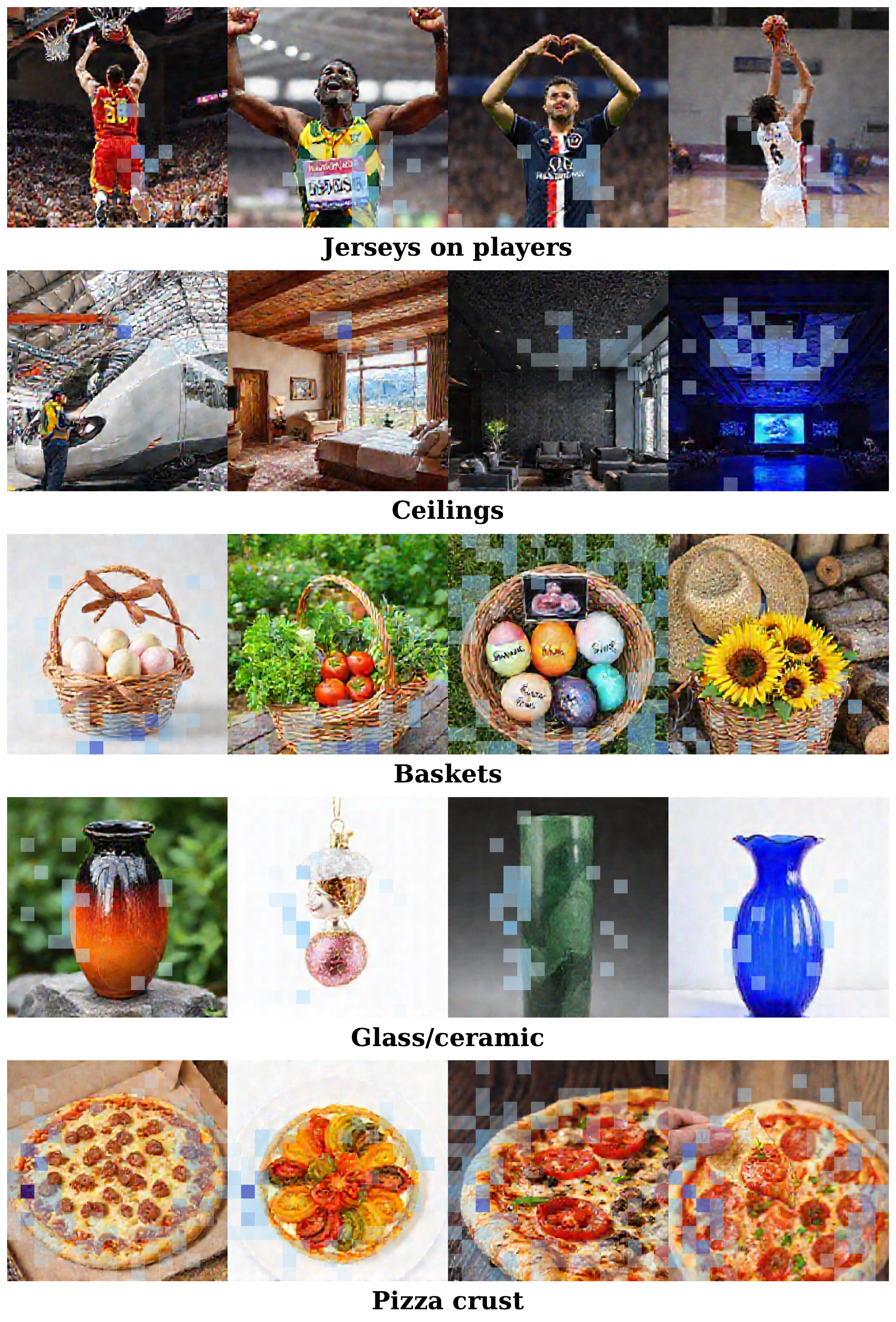}
    \caption{Maximum activating examples of some interpretable features from ITDA. }
    \label{fig:maxacts-itda}
\end{figure}

\Cref{fig:maxacts-more-saes} contains max activating examples for 5 different SAE features. \Cref{fig:maxacts-itda} contains max activating examples for 5 different ITDA features. We can notice a difference in pixel coverage with these two types of features.

\section{Specifics of the SAE training codebase}
\label{sec:impl}

We implemented our SAE training code in Jax for Google TPUs. To our knowledge, we produced the first Jax implementation of \flux. The diffusion transformer and the T5 text encoder take 48GB of memory combined at bfloat16 precision. We used v4-8 TPUs, which have 128GB of HBM combined. We implemented 4-bit NormalFloat quantization \cite{dettmers2023qloraefficientfinetuningquantized} together with inference kernels to avoid materializing the dequantized weight matrices in memory. We implemented FSDP across the output axis for partitioning the weights across devices.

We used Oryx and \texttt{jax.lax.cond\_clobber} to gather activations through multiple timesteps as outlined in \cite{kramar}. We had previously sent activations into CPU memory and found that the throughput was not reduced. It is possible that caching activations with the diffusion model is a bottleneck that overshadows a CPU-TPU transfer, or that we missed where the transfer occurs.

Finally, we sped up the TopK SAE decoder similarly to \cite{gao2024scaling}. We did not write a sparse matrix multiplication kernel for TPUs due to a lack of time, but we came up with a batched implementation that, while using up HBM, doesn't need to store all pre-activation values.

The overall most important practical improvements are running the decoder in vmap and collecting activations with layer-scanned Oryx. We share our TPU training code in \href{http://github.com/neverix/fae}{github.com/neverix/fae}.

\section{Autointerpretation prompts}
\label{sec:autointerp-prompt}

\begin{modelprompt}[title=Autointepretation prompt]
    You will be given a list of images.
Each image will have activations for a specific neuron highlighted in blue.
You should describe a common pattern or feature that the neuron is capturing.
First, write for each image, which parts are higlighted by the neuron.
Then, write a common pattern or feature that the neuron is capturing.
\end{modelprompt}

\begin{modelprompt}[title=Judge prompt]
    You will be given an image. And a neuron's activations description.
The image will have activations for the neuron highlighted in blue.
You should judge whether the description of the neuron's pattern is accurate or not.
Return a score between 0 and 1, where 1 means the description is accurate and 0 means it is not.
Be very critical. The pattern should be literal and specific, and vague or general descriptions should be rated low.
The activation pattern is \{pattern\}.
\end{modelprompt}

Our autointerpretation code is public at \href{https://github.com/kisate/flux-saes-gpu}{github.com/kisate/flux-saes-gpu}.

\end{document}